\documentclass[letterpaper,journal]{IEEEtran}

\usepackage{amsmath,amssymb}
\usepackage{booktabs}
\usepackage{cite}
\usepackage{graphicx}
\usepackage{mathtools}
\usepackage{multirow}
\usepackage{url}

\newcommand{\sr}{\mathrm{SR}}
\newcommand{\LN}{\operatorname{LN}}
\newcommand{\MHA}{\operatorname{MHA}}
\newcommand{\Swin}{\operatorname{Swin}}
\newcommand{\clip}{\operatorname{clip}}
\newcommand{\sg}{\operatorname{sg}}
\newcommand{\ind}[1]{\mathbb{I}\!\left[#1\right]}
\newcommand{\pos}[1]{\left[#1\right]_{+}}

\newcommand{\submissionnotice}{This work has been submitted to the IEEE for possible publication. Copyright may be transferred without notice, after which this version may no longer be accessible.}
\makeatletter
\def\@arxivnoticelines#1. #2\@nil{#1.\\#2}
\def\ps@arxivfirst{%
  \ps@plain
  \def\@oddhead{\hfil\smash{\parbox[b]{\textwidth}{\centering\normalfont\small\expandafter\@arxivnoticelines\submissionnotice\@nil}}\hfil}%
  \let\@evenhead\@oddhead}
\makeatother

\begin{document}

\title{Spatial Belief Tracking with Offline Vision-Language Distillation for Active Geo-Localization}

\author{Ziyang~Xu,~\IEEEmembership{Student Member,~IEEE,}
        Peng~Han, and~Hai~Zhou%
\thanks{Corresponding author: Peng Han.}%
\thanks{Ziyang Xu, Peng Han, and Hai Zhou are with the National Space
Science Center, Chinese Academy of Sciences, Beijing 100190, China.
E-mail: xuziyang23@mails.ucas.ac.cn, hanpeng@nssc.ac.cn, and zhouhai@nssc.ac.cn.}%
\thanks{Ziyang Xu is also with the University of Chinese Academy of Sciences,
Beijing 101408, China.}}

\maketitle
\thispagestyle{arxivfirst}

\begin{abstract}
Active geo-localization requires an agent to reach a target specified by an aerial image, a ground-level photograph, or a text description. The agent observes one aerial patch at a time and must choose where to look next while most of the search area remains unseen. We propose EAGLE, an agent that maintains a spatial belief over the search grid by predicting aerial features and estimating their uncertainty at each location. During offline training, a teacher uses vision-language features of the complete scene to guide the student's predictions at masked locations. During navigation, the student updates the grid from the aerial patches observed along its route. A convolutional policy chooses the next move using the similarity of each predicted feature to the goal, the estimated uncertainty at unvisited locations, and the visit history. An additional training reward favors moves that reduce total uncertainty over unvisited locations. Experiments show that a policy trained with aerial goals on MASA transfers to disaster imagery on xBD and can locate targets specified by ground-level photographs or text on MM-GAG without further policy training. We will release the code upon acceptance of the paper.
\end{abstract}


\section{Introduction}

\begin{figure}[!t]
\centering
\includegraphics[width=\columnwidth]{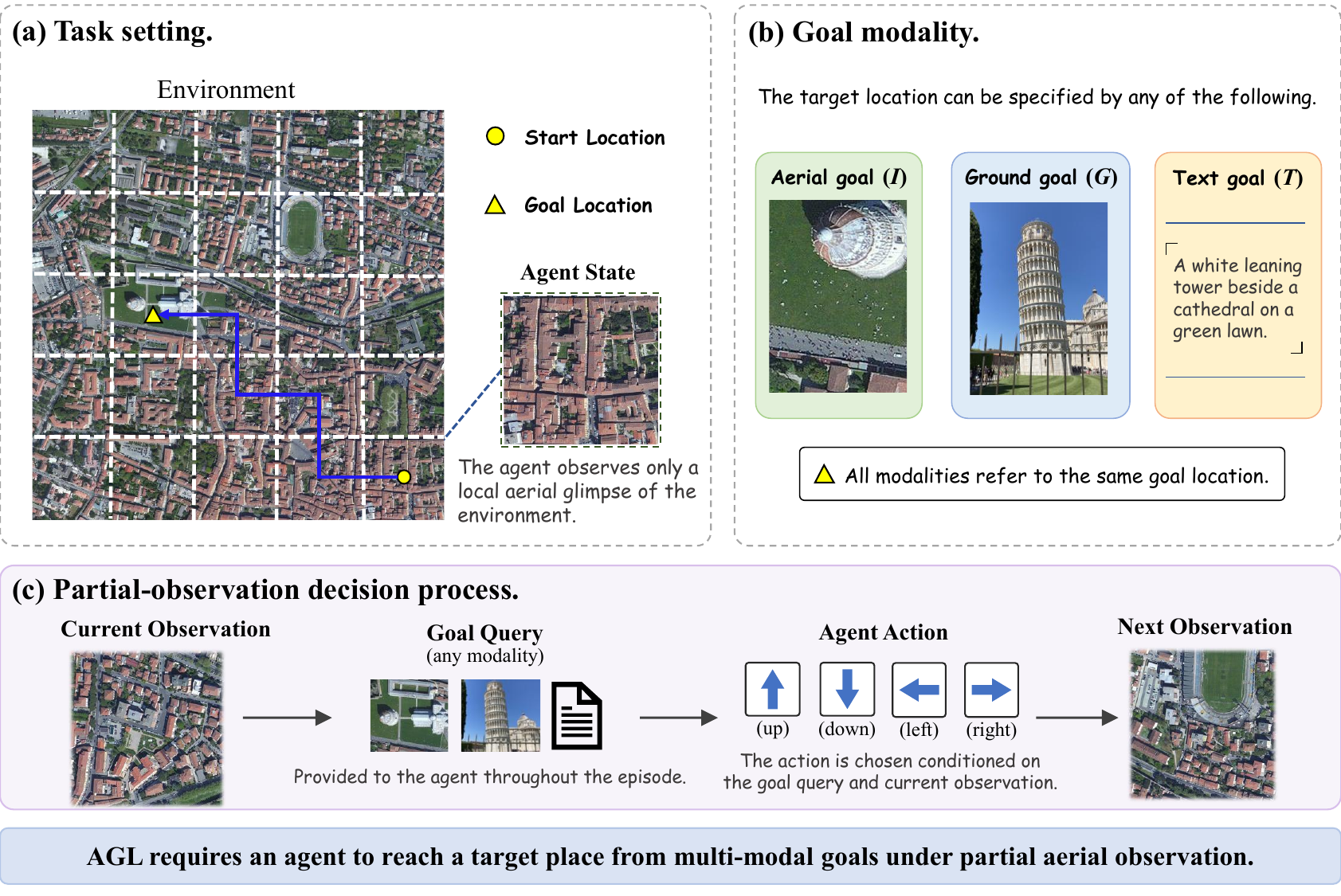}
\caption{Multi-modal active geo-localization. An aerial image, ground photograph, or text query identifies the same hidden target. The agent observes one local aerial patch at a time and uses four-neighbor moves to reach the target within a fixed budget.}
\label{fig:agl-task}
\end{figure}

\IEEEPARstart{S}{earch-and-rescue}, disaster assessment, and environmental monitoring often require an unmanned aerial vehicle to locate a target from indirect evidence such as an aerial crop, a ground photograph, or a natural-language description. Active geo-localization (AGL) formalizes this setting as goal-conditioned navigation over a discretized aerial search region~\cite{sarkar2024gomaa,mi2025geoexplorer}. Fig.~\ref{fig:agl-task} illustrates an episode. A fixed aerial, ground-image, or text query refers to a hidden target cell. The agent observes one local aerial patch at a time and selects one of four cardinal moves. It must reach the target within a fixed action budget. AGL therefore combines cross-view or language matching with exploration under partial observability. The agent must track evidence about unseen cells and allocate a short search trajectory efficiently.

GOMAA-Geo~\cite{sarkar2024gomaa} aligns aerial, ground, and text goals for a history-aware policy. GeoExplorer~\cite{mi2025geoexplorer} adds action-state dynamics and curiosity-driven exploration. Both methods condition their actions on the navigation history. A central difficulty is that most cells remain unseen when the agent chooses its next move. An agent that predicts features for those cells must also estimate how reliable each prediction is. A full-scene vision-language model can provide strong semantic supervision during training, but using it online would require privileged scene context and incur substantial inference cost. We therefore transfer its global semantic knowledge through offline distillation and deploy only a context-free tracker conditioned on accumulated partial observations.

\begin{figure*}[!t]
	\centering
	\includegraphics[width=0.98\textwidth]{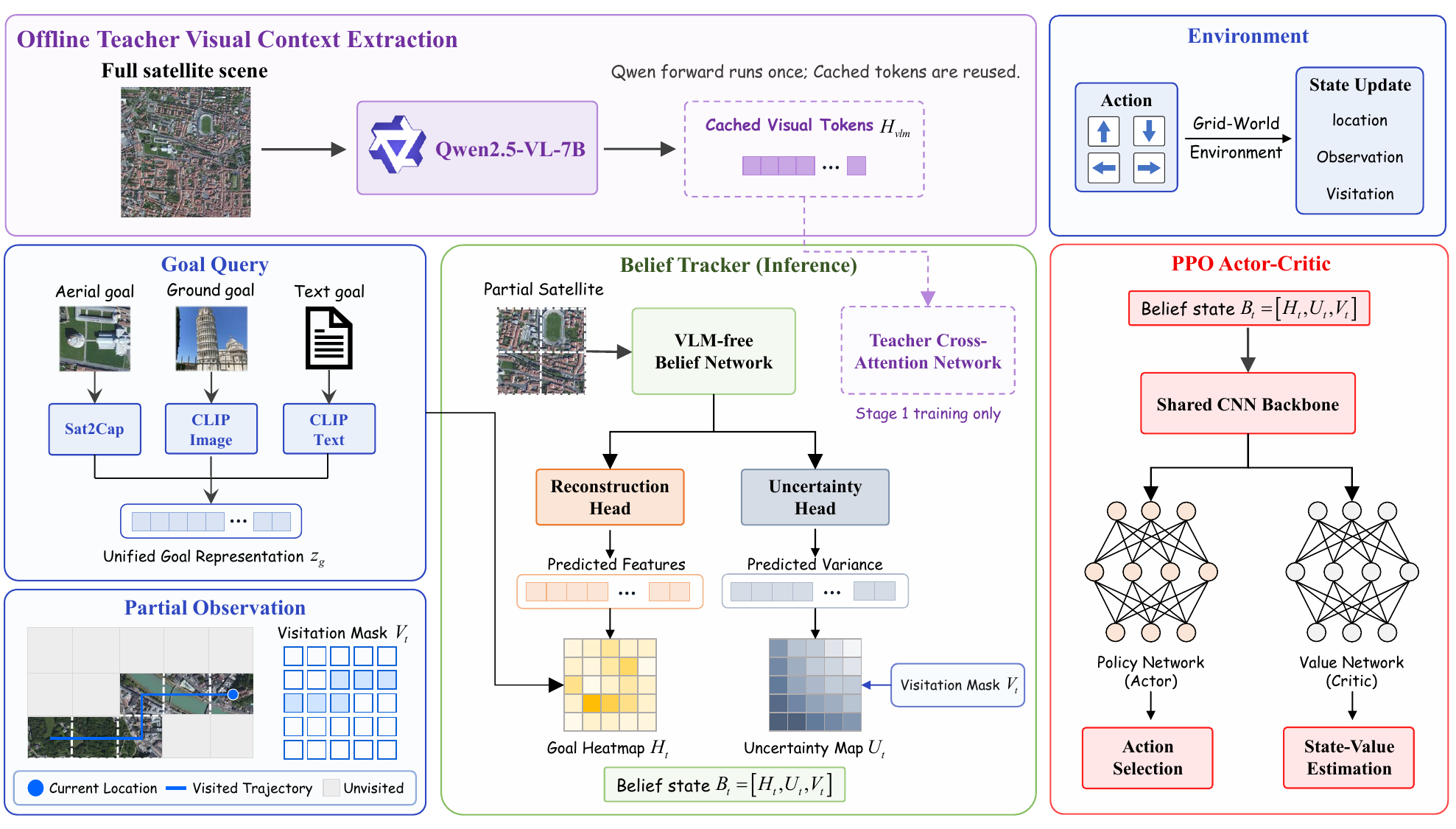}
	\caption{Overview of EAGLE. Offline VLM supervision trains a context-free tracker from partial observations. At deployment, predicted grid features and variance are combined with the aligned goal representation and trajectory-derived visitation map to form $B_t=[H_t,U_t,V_t]$. The PPO actor-critic uses this state without Qwen or full-scene context.}
	\label{fig:architecture}
\end{figure*}

We call our model EAGLE, short for Explicit Spatial-Belief Agent for Geo-Localization and Exploration. Its spatial belief records what the agent has observed and what it predicts at each grid location. As shown in Fig.~\ref{fig:architecture}, a tracker predicts an aerial feature and its variance for each cell from the partially observed mosaic. Matching those features to the goal produces a heatmap. The policy is a three-layer convolutional neural network. This CNN uses the heatmap, the estimated uncertainty, and the record of past visits to choose the next move. During offline training, the Qwen vision-language model supplies the teacher with full-scene context. At policy time, EAGLE uses the student tracker, the goal encoder, and the convolutional policy. The exploration bonus rewards reductions in uncertainty as the agent visits new cells.

Our contributions are as follows.
\begin{itemize}
	\item We introduce a grid-aligned spatial belief for multi-modal AGL that preserves cell coordinates and local neighborhoods in the policy state.
	
	\item We propose an offline distillation framework that transfers full-scene VLM semantics to a context-free tracker used at deployment.
	
	\item We develop a variance-reduction coverage objective that avoids privileged distance-based reward shaping.
\end{itemize}

\section{Related Work}

\subsection{Cross-View and Multimodal Alignment}
Cross-view geo-localization provides the representation foundation for matching an indirect query to aerial imagery. Early systems aligned ground and aerial views in a joint semantic space~\cite{workman2015wide}. VIGOR replaced ideal one-to-one pairing with dense aerial coverage and arbitrary query locations~\cite{zhu2021vigor}. Large-scale studies introduced scalable classification-based retrieval and standardized design comparisons~\cite{berton2022cosplace,berton2022deep}. Later work aligns images with continuous coordinates~\cite{vivanco2023geoclip}, improves tolerance to ground-view variation~\cite{mi2024congeo}, and expands evaluation to worldwide imagery~\cite{astruc2024osv5m}. Metric cross-view localization has also modeled position-dependent predictive uncertainty~\cite{fervers2023uncertainty}. These methods assume an available reference collection for passive retrieval or coordinate estimation. AGL reveals the aerial search region incrementally and must choose which cell to observe next.

Multimodal AGL also requires aerial, ground-level, and language goals to be comparable with observed aerial cells. CLIP established scalable image-text alignment and transferable visual representations~\cite{radford2021clip}. Sat2Cap learned an overhead encoder that predicts ground-level CLIP embeddings from paired overhead and ground imagery, connecting remote-sensing observations to the shared CLIP space~\cite{dhakal2024sat2cap}. EAGLE uses frozen, modality-aligned encoders to construct the goal embedding and aerial feature targets. Spatial belief updating operates on partial observations.

Multi-view retrieval offers a useful comparison. Chen et al.~\cite{chen2026dide} aggregate drone views into a representation for matching against satellite imagery. In an active search, the set of available views depends on earlier decisions. An incorrect match can therefore change the evidence available for every subsequent decision. EAGLE retains the location of each observation and compares a goal against a feature grid that changes as observations arrive. In aerial vision-language navigation, an instruction can also specify a sequence of movements and landmarks along a route~\cite{xu2026aerial}. Here the query describes a destination. The agent chooses its own route and receives the same query at every step.

\subsection{VLM Teachers and Knowledge Distillation}
Large vision-language models such as Qwen2.5-VL, LLaVA, and InternVL support broad visual-language understanding and reasoning~\cite{bai2025qwen25vl,liu2023llava,chen2024internvl}, but invoking them at every step would add substantial cost to reinforcement-learning rollouts. Knowledge distillation transfers representations or predictions from a high-capacity teacher to a smaller task-specific student~\cite{hinton2015distilling}. In vision-language learning, DistillVLM aligns visual tokens and transfers intermediate representations~\cite{fang2021distillvlm}, TinyCLIP preserves cross-modal alignment through affinity matching and weight inheritance~\cite{wu2023tinyclip}, and EfficientVLM combines distillation with modal-adaptive pruning~\cite{wang2023efficientvlm}. These methods primarily compress pretrained models or preserve their learned multimodal representations.

EAGLE instead uses full-scene VLM context as privileged supervision for partial-observation inference. Its teacher-conditioned branch combines the partial mosaic with context extracted from the complete aerial scene, whereas the student receives only the same partial mosaic. After distillation, the VLM and teacher branch are removed, leaving a context-free tracker for deployment.

\subsection{Active Geo-Localization and Exploration}
AiRLoc formulated active aerial localization as reinforcement learning over a partially observed search grid~\cite{pirinen2022aerial}. GOMAA-Geo extended the task to aerial, ground-image, and language goals through contrastive representation learning and history-aware navigation~\cite{sarkar2024gomaa}. GeoExplorer coupled action-state modeling with curiosity-driven exploration to improve generalization~\cite{mi2025geoexplorer}. Decision Transformer models trajectories as sequences for offline control~\cite{chen2021decision}. Related vision-and-language navigation work studies instruction-conditioned embodied agents~\cite{gu2022vlnsurvey}, continuous drone flight~\cite{liu2023aerialvln}, and reusable visual navigation policies across robot embodiments~\cite{shah2023vint}. AGL differs because the query identifies a hidden target, and each observation reveals one selected aerial cell.

Partial observability motivates belief-state estimation~\cite{kaelbling1998pomdp,thrun2002probabilistic} and spatial memory~\cite{gupta2017cognitive,chaplot2020activeslam}. Intrinsic-motivation methods reward predictive surprise~\cite{stadie2015incentivizing,pathak2017curiosity}, evaluate curiosity at scale~\cite{burda2019largescale}, distill novelty through a fixed random target network~\cite{burda2019exploration}, or reward progress in learning a world model~\cite{kim2020progress}. DynCur-Geo~\cite{sun2026dyncur} varies the weight on its prediction-error reward with the remaining goal distance. EAGLE applies heteroscedastic uncertainty~\cite{nix1994variance} to spatial belief tracking and exploration in multi-modal AGL.

Predicting beyond the observed map has also been studied in indoor object search. Georgakis et al.~\cite{georgakis2022} infer semantic categories outside the field of view and use uncertainty to choose exploration goals. Their map can be queried by object category. A ground photograph of a particular building, however, can distinguish among locations that all belong to the same category. EAGLE predicts a continuous aerial feature at each location so that one map can be queried by an individual image or description. This choice also defines the uncertainty target. The residual is measured in the aerial embedding space used to match the query at deployment.

\section{Method}

\subsection{System Overview}
Following GOMAA-Geo and GeoExplorer, the search area is a $5\times5$ grid of non-overlapping aerial patches. Let $s_t$ and $s_g$ denote the current cell and hidden goal cell. The action set is $\{\mathrm{up},\mathrm{down},\mathrm{left},\mathrm{right}\}$. Frozen modality-specific encoders produce the 512-dimensional goal embedding $z_g\in\mathbb{R}^{512}$. Aerial crops use the benchmark's Sat2Cap aerial encoder, ground photographs use the CLIP ViT-B/32 vision tower, and text uses the corresponding CLIP text tower~\cite{dhakal2024sat2cap,radford2021clip}. These encoders remain outside the EAGLE optimization graph. An episode succeeds when the agent reaches $s_g$ within $T_{\max}=10$ moves. Evaluation reports success ratio $\sr$ at Manhattan start-goal distances $C\in\{4,5,6,7,8\}$.

\begin{figure*}[!t]
	\centering
	\includegraphics[width=0.98\textwidth]{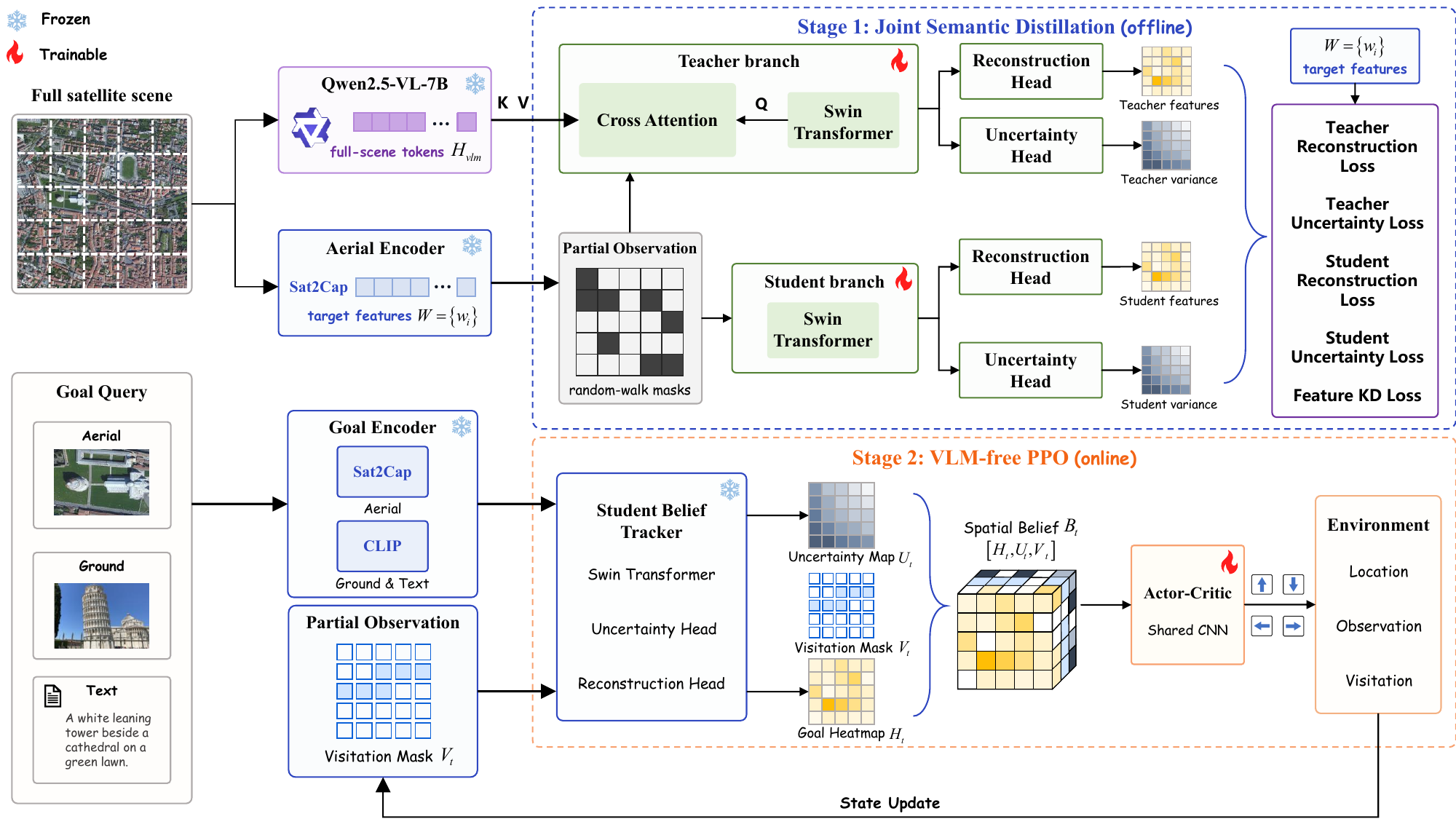}
	\caption{Two-stage training. Stage~1 jointly optimizes a Qwen-conditioned teacher and a context-free student from identical random-walk partial mosaics. Both branches receive masked-cell aerial-feature reconstruction and branch-specific heteroscedastic supervision. The student also matches detached teacher features. Stage~2 removes the teacher path and trains the CNN policy with a frozen student tracker.}
	\label{fig:two-stage}
\end{figure*}

EAGLE follows the two-stage pipeline shown in Figs.~\ref{fig:architecture} and~\ref{fig:two-stage}. Stage~1 jointly trains a full-scene VLM-conditioned teacher and a context-free student on the same randomly masked partial mosaics. Both branches infer cell-level aerial features and predictive variance, but only the teacher receives frozen full-scene VLM context. At policy time, the student features are matched with the goal embedding to produce a goal heatmap, while its predictive variance is gated by trajectory visitation to produce an uncertainty map. These maps and the trajectory-derived visitation map form the grid-aligned state $B_t=[H_t,U_t,V_t]$. In Stage~2, we freeze the student tracker and train a convolutional policy with proximal policy optimization. The PPO policy acts on $B_t$. Inference retains the frozen modality-aligned goal encoder, student tracker, and CNN policy.

\subsection{Building the Observation History}
At reset, the environment reveals the RGB patch at the starting cell. A move reveals the patch at the new grid coordinate and sets the corresponding entry in the visitation map to one. We place every revealed patch at its original coordinate in a common canvas. For the standard aerial patches each cell occupies $300\times300$ pixels, giving a $1500\times1500$ canvas. Unobserved cells contain zeros. A repeated visit preserves the previously observed patch and its grid coordinate.

Three quantities refer to different aspects of the episode. The goal query is fixed throughout the search. The observed canvas grows whenever the agent reaches a new cell. The current location can change on every move, including moves between two cells that have already been observed. Only the last of these quantities determines which neighboring cells can be reached on the next move.

The benchmark feature grid gives the tracker a target at every spatial coordinate during offline training. Its entries are frozen aerial embeddings. They provide a common output space for scenes with very different appearances. Two views may differ in color or texture yet be close in the embedding used to describe the goal. At test time the policy needs one embedding for the goal and a prediction of the grid from its observations. It can score a candidate cell before seeing its aerial patch. When the patch is subsequently revealed, the tracker computes the grid again using the additional pixels.

\subsection{Dual-Branch Spatial Belief Tracker}

\paragraph{Offline teacher context}
We stitch the 25 aerial patches into a complete scene, resize it to $896\times896$ pixels, and run frozen Qwen2.5-VL-7B once per training scene. We take the visual tokens between the image delimiters in the final hidden layer. For sequences longer than 1,024 tokens, we retain 1,024 evenly spaced entries in their original order. We pad shorter sequences with zeros. A fixed linear projection with orthogonally initialized weights maps each token from 3,584 to 1,024 dimensions. We cache the projected tokens in half precision as $H_{\mathrm{vlm}}$ for the teacher.

\paragraph{Partial-observation tracker}
At time $t$, the observation mask $O_t\in\{0,1\}^{5\times5}$ constructs the stitched mosaic $I_t$. Cells with $O_t(i)=0$ are blank. The teacher and student use separate Swin backbones~\cite{liu2021swin}. Both encode the same masked mosaic $I_t$. Let $b\in\{T,S\}$ index the teacher and student branches, and define $\delta_b\triangleq\ind{b=T}$. The branch encoder is
\begin{equation}
\begin{aligned}
F_t^{\,b}&=\Swin_b(I_t),\\
\widetilde F_t^{\,b}
&=\LN\!\left(
F_t^{\,b}
+\delta_b\MHA\!\left(
F_t^{\,b},H_{\mathrm{vlm}},H_{\mathrm{vlm}}
\right)\right).
\end{aligned}
\label{eq:dual-branch-encoder}
\end{equation}
In the teacher branch, the Swin tokens from the partial mosaic supply the queries for eight-head cross-attention. The cached Qwen tokens supply both keys and values. The attention layer uses the positional information already encoded by the two backbones, with no additional position embedding. The student processes the partial mosaic with its own Swin encoder.

Separate reconstruction and uncertainty heads $\mathcal R_b$ and $\mathcal U_b$ produce one feature vector and one scalar log-variance for each cell.
\begin{equation}
\begin{aligned}
\widehat W_t^{\,b}
&=\mathcal R_b(\widetilde F_t^{\,b})
\in\mathbb R^{N\times d},\\
\boldsymbol\ell_t^{\,b}
&=\mathcal U_b(\widetilde F_t^{\,b})
\in\mathbb R^N,\\
\overline{\boldsymbol\sigma}_t^{\,2,b}
&=\exp\!\left[
\clip\!\left(
\boldsymbol\ell_t^{\,b};
\log\sigma_{\min}^{2},
\log\sigma_{\max}^{2}
\right)\right].
\end{aligned}
\label{eq:grid-predictions}
\end{equation}
Here $N=25$ is the number of cells in the $5\times5$ search grid, and $d=512$ is the dimensionality of each aerial feature embedding. The constants $\sigma_{\min}^{2}=10^{-4}$ and $\sigma_{\max}^{2}=64$ are the lower and upper bounds on the predicted variance, respectively. Exponentiation and clipping are element-wise. The $i$-th row $\widehat{\boldsymbol w}_{t,i}^{\,b}\in\mathbb R^d$ is the predicted aerial feature of cell $i$. The scalar $\overline{\sigma}_{t,i}^{\,2,b}$ is shared across its $d$ feature dimensions.

Only the context-free student is retained after Stage~1. For grid coordinates $r,c\in\{1,\ldots,5\}$, let $\iota(r,c)=5(r-1)+c$ denote the row-major cell index. At policy time, the goal heatmap and uncertainty map are
\begin{equation}
\begin{aligned}
H_t(r,c)
&=
\frac{
\left(\widehat{\boldsymbol w}_{t,\iota(r,c)}^{\,S}\right)^\top
\boldsymbol z_g
}{
\left\|\widehat{\boldsymbol w}_{t,\iota(r,c)}^{\,S}\right\|_2
\left\|\boldsymbol z_g\right\|_2
},\\
U_t(r,c)
&=
\left(1-V_t(r,c)\right)
\overline{\sigma}_{t,\iota(r,c)}^{\,2,S},\\
B_t
&=
\left[H_t\;U_t\;V_t\right]^\top
\in\mathbb R^{3\times5\times5}.
\end{aligned}
\label{eq:spatial-belief}
\end{equation}
The goal enters the CNN through $H_t$. Each cell contains the cosine similarity between the goal embedding and the aerial feature the student predicts for that cell. A text description thus reaches the policy as 25 similarity scores. The other two input channels are $U_t$, the predicted variance with visited cells set to zero, and $V_t$, the cumulative map of visited cells. The environment tracks the current cell, updates $V_t$ after each move, and masks out-of-bounds actions at the actor output before action selection. The CNN receives no separate current-position channel.

\paragraph{Spatial resolution of the tracker}
Each branch uses a Swin-Base backbone initialized with ImageNet weights. Its last spatial feature map has 1,024 channels. On the standard canvas the map is $47\times47$, so the teacher has 2,209 spatial queries for cross-attention. A query can draw on any of the cached scene tokens. The number of queries depends on the Swin feature map, while the fixed length of the cache determines the number of keys and values. The attention output is added at the corresponding query location. Both branches restore the spatial layout before predicting the grid.

The reconstruction and uncertainty heads each apply adaptive average pooling to a $5\times5$ output. A pointwise convolution in the reconstruction head maps the pooled channels to a 512-dimensional feature for each cell. A separate pointwise convolution in the uncertainty head returns a scalar at the same location. This ordering matters for an aerial search. The cell index used to score the goal must identify the same patch that is marked on the visit map. Row-major indexing is used in the mosaic, the reconstruction target, and the policy input.

\paragraph{Meaning of the uncertainty estimate}
The variance has units of squared aerial-feature residual per feature dimension. It is learned separately from the cosine score used for the goal. For a fixed observation history, changing the query can move the peak in $H_t$ to another part of the grid while the raw variance prediction is identical. Conversely, revealing an additional patch can affect both predictions, including those for nearby cells still outside the observed region. Once a cell has been seen, the visit map gates its contribution to the coverage reward to zero. The underlying feature estimate for that cell still participates in the goal comparison.

\subsection{Stage 1: Dual-Branch Semantic Distillation}
For each training scene we draw a masked fraction $r$ uniformly between 0.1 and 0.9. We round $25(1-r)$ to the nearest integer to set the number of distinct cells to reveal, with a minimum of one. Starting from a uniformly sampled cell, a random walk chooses one of the four directions with equal probability at each step. It stays in place at a blocked boundary and stops once it has visited the target number of distinct cells. Both branches see the same visible cells. For Stage~1 validation we draw a fresh binary mask for each scene in each epoch. Let $\mathcal M_t$ denote the masked cells and angle brackets the mean over those cells.
\begin{equation*}
\begin{aligned}
\mathcal M_t
&\triangleq
\{i\in\{1,\ldots,N\}:O_t(i)=0\},\\
\left\langle a_i\right\rangle_{i\in\mathcal M_t}
&\triangleq
\frac{1}{|\mathcal M_t|}
\sum_{i\in\mathcal M_t}a_i.
\end{aligned}
\end{equation*}
Let $W=\{\boldsymbol w_i\}_{i=1}^{N}$ be the frozen benchmark aerial-feature grid. For branch $b\in\{T,S\}$, the dimension-averaged cell residual is
\begin{equation}
e_{t,i}^{\,b}
=
\frac{1}{d}
\left\|
\widehat{\boldsymbol w}_{t,i}^{\,b}
-\boldsymbol w_i
\right\|_2^2,
\qquad
i\in\mathcal M_t,\quad b\in\{T,S\}.
\label{eq:cell-residual}
\end{equation}
Let $\sg[\cdot]$ denote the stop-gradient operator, which acts as
the identity in the forward pass and has zero gradient in the backward pass. Each branch uses separate reconstruction and detached-residual heteroscedastic objectives.
\begin{equation}
\begin{aligned}
\mathcal L_{\mathrm{rec}}^{\,b}
&=
\left\langle e_{t,i}^{\,b}\right\rangle_{i\in\mathcal M_t},\\
\mathcal L_{\mathrm{nll}}^{\,b}
&=
\left\langle
\frac{\sg[e_{t,i}^{\,b}]}
{2\overline{\sigma}_{t,i}^{\,2,b}}
+\frac{1}{2}\log
\overline{\sigma}_{t,i}^{\,2,b}
\right\rangle_{i\in\mathcal M_t},\\
\mathcal L_b
&=
\mathcal L_{\mathrm{rec}}^{\,b}
+\lambda_u\mathcal L_{\mathrm{nll}}^{\,b},
\qquad
\lambda_u=0.005,\quad b\in\{T,S\}.
\end{aligned}
\label{eq:branch-objective}
\end{equation}
The variance loss treats the reconstruction residual as a fixed target. Its gradient passes through the uncertainty head into the branch encoder. As the reconstruction head shares this encoder, these updates can change its predictions on subsequent forward passes. On masked cells, the student also learns to match the teacher's feature predictions.
\begin{equation}
\mathcal L_{\mathrm{KD}}
=
\frac{1}{d}
\left\langle
\left\|
\widehat{\boldsymbol w}_{t,i}^{\,S}
-\sg[\widehat{\boldsymbol w}_{t,i}^{\,T}]
\right\|_2^2
\right\rangle_{i\in\mathcal M_t}.
\label{eq:feature-distillation}
\end{equation}
The detached teacher prediction makes this term update the student branch alone. The complete Stage~1 objective is
\begin{equation}
\mathcal L_{\mathrm{Stage1}}
=
\sum_{b\in\{T,S\}}\mathcal L_b
+\lambda_{\mathrm{kd}}\mathcal L_{\mathrm{KD}}.
\label{eq:stage1-objective}
\end{equation}
We set $\lambda_{\mathrm{kd}}=0.5$ and evaluate each loss on the masked cells. Only $\mathcal L_T$ updates the teacher.

\paragraph{Interpreting residual supervision}
For a masked cell with fixed reconstruction residual $e>0$, write the predicted variance as $v$. Its contribution to the uncertainty objective is $e/(2v)+\log v/2$. Differentiating with respect to $\log v$ gives $(1-e/v)/2$. When $e$ exceeds $v$, the gradient increases the fitted variance under gradient descent. When $e$ falls below $v$, the gradient acts in the opposite direction. For a set of residuals that share one prediction, the minimizing variance is their arithmetic mean, subject to the clipping bounds. This explains why a single scalar can summarize the expected feature error across dimensions at one grid cell.

A high cosine score and a large residual variance are compatible. For example, multiplying a nonzero feature vector by a positive scalar preserves its cosine similarity with every possible goal, but changes its squared reconstruction error against the frozen aerial target. The policy receives both quantities so that the reliability signal need not be inferred from the magnitude of the goal score. Keeping the target residual detached during optimization preserves this interpretation of the variance term.

\subsection{Stage 2: Belief-Space Policy Learning}
After distillation, Qwen, $H_{\mathrm{vlm}}$, and the teacher-conditioned branch are removed. The context-free student tracker is frozen and receives the accumulated partial mosaic. The CNN policy is trained with PPO~\cite{schulman2017ppo} and GAE~\cite{schulman2015high}. Let $m_t$ denote the highest heatmap score encountered along the trajectory up to time $t$. It is initialized at the starting cell as

\begin{equation*}
m_0=H_0(s_0).
\end{equation*}
For each transition arriving at time $t\ge1$, the recursive score and reward are
\begin{equation}
\begin{aligned}
\Delta H_t
&=H_t(s_t)-m_{t-1},\\
m_t
&=\max\{m_{t-1},H_t(s_t)\},\\
R_t
&=-c_{\mathrm{step}}
+\alpha_H\ind{\Delta H_t>0}
+\alpha_G\ind{s_t=s_g}\\
&\quad
+\lambda_{\mathrm{int}}
\pos{\lVert U_{t-1}\rVert_1-\lVert U_t\rVert_1},
\qquad t\ge1.
\end{aligned}
\label{eq:policy-reward}
\end{equation}
Here $\mathbb{I}[\cdot]$ is the indicator function and
$[x]_+=\max(x,0)$ denotes the positive part. Here $\lVert U_t\rVert_1$ sums the nonnegative entries over all grid cells. We set the step cost
$c_{\mathrm{step}}=0.01$, the heatmap-improvement bonus
$\alpha_H=0.20$, the terminal success bonus $\alpha_G=2.0$, and the
variance-reduction weight $\lambda_{\mathrm{int}}=0.5$.

The heatmap term rewards a transition when the current cell attains a
higher predicted goal-agreement score than any score previously encountered along the trajectory. The variance-reduction term rewards transitions that decrease the aggregate uncertainty represented by the trajectory-gated map $U_t$. These nonterminal shaping terms are computed entirely from the student's policy-time belief and do not use hidden-goal distance. The goal coordinate $s_g$ is used only by the terminal success condition.

\begin{table}[!t]
\caption{One navigation step with the frozen student. The order applies to each of the three goal modalities.}
\label{tab:online-step}
\centering\small
\begin{tabular}{@{}cl@{}}\toprule
Step & Operation \\ \midrule
1 & Write the observed patch at the current grid cell.\\
2 & Update the visit mask and form the partial mosaic.\\
3 & Predict the aerial feature and variance of every cell.\\
4 & Compare cell features with the fixed goal vector.\\
5 & Stack similarity, gated variance, and visit history.\\
6 & Evaluate the actor and mask out-of-bounds moves.\\
7 & Execute a legal move and test for termination.\\
\bottomrule\end{tabular}
\end{table}

\subsection{Optimization and Action Selection}
\paragraph{Training the tracker}
The complete image used by Qwen is accompanied by a fixed request to describe the geography and terrain. The input is processed in one forward pass. We read hidden states at visual-token positions and cache the projected array, so the same scene context can be reused with different random walks. A short walk and a long walk through the same scene can present very different reconstruction problems while sharing the same cache and aerial-feature targets. Each mask determines which of those target cells contribute to the loss.

We optimize the independent teacher and student with AdamW at a constant learning rate of $10^{-4}$ and weight decay of $10^{-4}$. The training microbatch contains a single scene and gradients accumulate over eight microbatches, with an update on the final partial group. The checkpoint criterion is the mean joint objective on the validation split. The configuration allows up to 1,000 epochs, with patience of 50 epochs and a minimum improvement of $10^{-4}$. Validation masks are drawn independently at each cell with equal probabilities of observing and masking it. A degenerate mask is adjusted to include at least one cell of each kind.

\paragraph{From a belief grid to four moves}
The policy first applies three $3\times3$ convolutions with 32, 64, and 128 output channels and rectified linear activations. Padding preserves the grid dimensions. Flattening the output gives 3,200 values. Separate actor and critic heads each have a 256-unit hidden layer. The actor returns one logit per cardinal move and the critic predicts a scalar value. An invalid move receives zero probability after masking. The critic evaluates the grid before action masking. Together the two heads and their shared convolutional layers contain 1,733,445 trainable parameters during policy learning.

PPO samples complete episodes from the MASA training split. It draws a start-to-goal distance uniformly from one through eight and uses the aerial embedding of the goal patch as its query. Training samples a legal move from the categorical policy. For evaluation we use the legal move with the highest logit. Thus the observation update, feature extraction, and boundary handling are the same for a sampled training action and a greedy evaluation action.

Table~\ref{tab:online-step} summarizes the recurring computations. The goal encoder runs before the first step. Only the visit history and observed pixels need to be carried forward to reconstruct the next belief. During training, the environment also retains the previous total uncertainty and the best heatmap score encountered along the trajectory to compute the reward. During greedy evaluation, the policy can choose its action from the current maps and boundary mask alone.

\begin{table*}[!t]
	\caption{Main results. Success ratio (\%) by Manhattan start-goal
		distance and dataset. For MM-GAG, the aerial-goal block includes the
		full aerial-goal comparison after omitting Random. The ground-image
		and text blocks compare EAGLE with GOMAA-Geo and GeoExplorer.}
	\label{tab:main}
	\centering
	\small
	\setlength{\tabcolsep}{3.5pt}
	\renewcommand{\arraystretch}{0.92}
	
	\begin{tabular*}{0.99 \textwidth}{
			@{\extracolsep{\fill}}
			c c c c c c c c
			@{}
		}
		\toprule
		Dataset & Method & $C=4$ & $C=5$ & $C=6$ & $C=7$ & $C=8$ & Avg.   \\
		\midrule
		
		\multirow{7}{*}{MASA}
		& Random & 14.1 & 5.8 & 6.4 & 2.5 & 2.4 & 6.2 \\
		& PPO & 14.3 & 17.8 & 19.2 & 22.7 & 26.0 & 20.0 \\
		& AiRLoc & 17.9 & 15.6 & 21.3 & 24.2 & 23.9 & 20.6 \\
		& DiT & 20.1 & 29.6 & 35.7 & 42.2 & 45.6 & 34.6 \\
		& GOMAA-Geo & 40.9 & 50.6 & 71.7 & 80.3 & 78.5 & 64.4 \\
		& GeoExplorer & 43.2 & 53.2 & 81.6 & 92.3 & 95.0 & 73.0 \\
		& \textbf{EAGLE}
		& \textbf{47.3} & \textbf{56.4} & \textbf{87.0}
		& \textbf{99.6} & \textbf{100.0} & \textbf{78.1} \\
		\midrule
		\multirow{7}{*}{xBD} & Random & 14.1 & 5.8 & 6.4 & 2.5 & 2.4 & 6.2 \\ & PPO & 11.3 & 11.5 & 12.9 & 16.7 & 19.5 & 14.4 \\ & AiRLoc & 12.0 & 13.0 & 15.1 & 16.3 & 19.9 & 15.3 \\ & DiT & 10.1 & 23.9 & 30.7 & 33.9 & 35.4 & 26.8 \\ & GOMAA-Geo & 40.0 & 46.3 & 65.5 & 73.9 & 69.4 & 59.0 \\ & GeoExplorer & 39.8 & 50.3 & 71.9 & 81.9 & 79.2 & 64.6 \\ & \textbf{EAGLE} & \textbf{44.0} & \textbf{54.2} & \textbf{85.4} & \textbf{98.6} & \textbf{100.0} & \textbf{76.4} \\
		\midrule
		\multirow{6}{*}{\shortstack{MM-GAG\\(Aerial)}}
		& PPO & 14.9 & 18.5 & 18.8 & 21.8 & 24.3 & 19.7 \\
		& AiRLoc & 17.5 & 16.9 & 20.2 & 21.6 & 22.9 & 19.8 \\
		& DiT & 20.2 & 28.6 & 35.2 & 41.9 & 44.2 & 34.0 \\
		& GOMAA-Geo & 40.9 & 50.6 & 66.4 & 73.6 & 70.2 & 60.3 \\
		& GeoExplorer & 43.4 & \textbf{54.2} & 76.3 & 83.7 & 82.8 & 68.1 \\
		& \textbf{EAGLE} & \textbf{45.9} & 52.9 & \textbf{86.8} & \textbf{98.2} & \textbf{100.0} & \textbf{76.8} \\
		\midrule
		\multirow{3}{*}{\shortstack{MM-GAG\\(Ground)}}
		& GOMAA-Geo & 43.8 & \textbf{51.5} & 68.1 & 74.9 & 68.9 & 61.4 \\
		& GeoExplorer & 43.1 & 51.4 & 72.0 & 82.5 & 78.2 & 65.4 \\
		& \textbf{EAGLE} & \textbf{47.1} & 51.4 & \textbf{85.2} & \textbf{97.9} & \textbf{100.0} & \textbf{76.3} \\
		\midrule
		\multirow{3}{*}{\shortstack{MM-GAG\\(Text)}}
		& GOMAA-Geo & 40.0 & 49.8 & 67.7 & 77.0 & 66.0 & 60.1 \\
		& GeoExplorer & 44.3 & 48.9 & 72.0 & 80.6 & 76.3 & 64.4 \\
		& \textbf{EAGLE} & \textbf{44.9} & \textbf{51.1} & \textbf{86.2} & \textbf{100.0} & \textbf{100.0} & \textbf{76.4} \\
		\bottomrule
	\end{tabular*}
	
\end{table*}

\paragraph{Updating the policy}
A rollout buffer holds all transitions from 512 episodes. With a ten-action budget, a buffer contains at most 5,120 transitions. Episodes that reach the goal early contribute fewer steps. Each transition stores its belief grid, legal-action mask, sampled action, log probability, reward, terminal flag, and value estimate. Preserving the action mask also preserves the support of the categorical distribution when that transition is used for a policy update.

Let $p_t$ denote the probability of the recorded action under the updated policy divided by its probability during collection, and let $\widehat A_t$ denote its normalized advantage. For a minibatch of transitions the actor maximizes
\begin{equation}
\mathbb E_t\!\left[\min\!\left\{p_t\widehat A_t,
 \clip[p_t,1-\epsilon,1+\epsilon]\widehat A_t\right\}\right].
\label{eq:ppo-update}
\end{equation}
We estimate advantages over the complete buffer with discount 0.99 and generalized advantage parameter 0.95. A terminal transition resets the recursion before the next episode. The critic minimizes the squared error against the return target. We use a value-loss weight of 0.5 and an entropy bonus of 0.01. Each of four passes shuffles the buffer and partitions it into minibatches of at most 1,024 transitions. Both heads share an AdamW update with learning rate $2\times10^{-4}$, clipping radius $\epsilon=0.2$, and maximum gradient norm 0.5. The student remains frozen throughout these updates.

\section{Experiments}

\subsection{Experimental Setup}

\paragraph{Datasets}
We train EAGLE on MASA with aerial goals and evaluate it on the held-out MASA split. We use xBD~\cite{gupta2019xbd} to test transfer to disaster imagery and MM-GAG~\cite{sarkar2024gomaa} to test aerial, ground-image, and text goals with the policy trained on MASA.

\begin{figure*}[!t]
	\centering
	\includegraphics[width=0.98\textwidth]{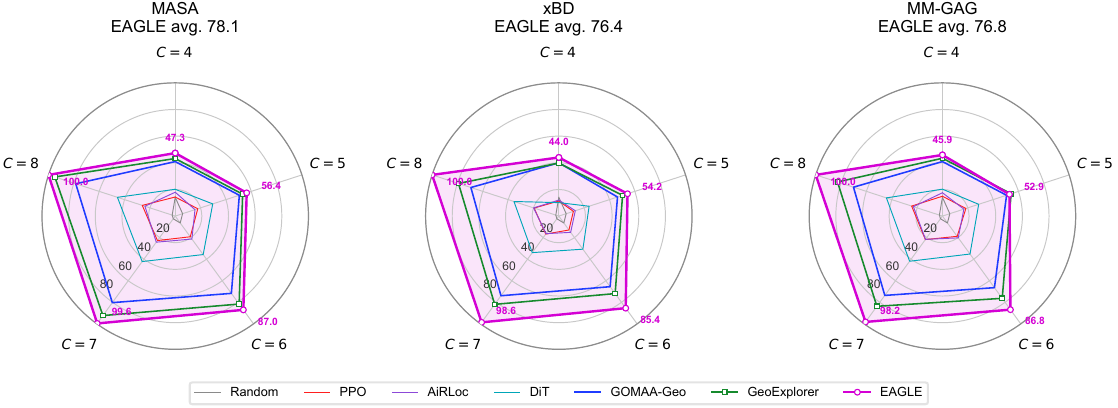}
	\caption{Distance-wise success ratios on MASA, xBD, and the MM-GAG aerial-goal split. Numeric labels mark EAGLE. Table~\ref{tab:main} reports exact values and the additional MM-GAG goal modalities. Baseline curves show previously reported success ratios.}
	\label{fig:results}
\end{figure*}

\paragraph{Navigation settings}
We follow GOMAA-Geo~\cite{sarkar2024gomaa} and GeoExplorer~\cite{mi2025geoexplorer} in the choice of grid size, action space, search budget, and start-to-goal distances. We compute EAGLE's average success ratio as the unweighted mean across the five distances.

\paragraph{Constructing cross-modal episodes}
Each MM-GAG search is cropped from a larger $11\times11$ array of aerial tiles. The ground photograph and text describe the location of the central aerial tile in this larger array. To place that location at a sampled goal cell in the $5\times5$ episode, we translate the crop window. For example, placing the central tile at the upper left of the search window makes the rest of the window extend to its right and below it. The aerial query uses the embedding of that same central tile. All three queries therefore identify one physical location even as its index within the search window varies.

For a given distance, the evaluation procedure reuses the same sampled start and goal coordinates across EAGLE's three modalities. At reset the input mosaic and visit mask for a particular configuration are consequently the same across modalities. The first decision is conditioned on a different heatmap when the goal encoder changes. Later decisions can also depend on differences in the observed imagery as the paths diverge. An episode contributes one binary outcome, reaching the specified cell within the budget, regardless of how many other cells the policy has visited.

\subsection{Evaluation of EAGLE}

\paragraph{In-Domain Evaluation}
On the held-out MASA scenes, EAGLE has the highest reported success ratio at each distance in Table~\ref{tab:main} and averages 78.1\%. At $C=4$, it reaches the goal in 47.3\% of episodes, compared with 43.2\% for GeoExplorer. At $C=7$, the corresponding values are 99.6\% and 92.3\%, as shown in Fig.~\ref{fig:results}.

\paragraph{Zero-Shot Cross-Domain Transfer}
EAGLE achieves a mean success ratio of 76.4\% on xBD and 76.8\% with aerial goals on MM-GAG. The published GeoExplorer averages are 64.6\% and 68.1\%, respectively. On the MM-GAG aerial task, GeoExplorer scores higher at $C=5$, while EAGLE records its largest margins at distances of six cells or more.

For both datasets the frozen student predicts features from the observed aerial patches. The policy uses the resulting maps with the same weights learned on MASA.

\paragraph{Zero-Shot Cross-Modal Generalization}
PPO training uses only aerial goals. On MM-GAG, the same policy achieves a mean success ratio of 76.8\% with aerial goals, 76.3\% with ground photographs, and 76.4\% with text. GeoExplorer reports 68.1\%, 65.4\%, and 64.4\% in the same order. The three EAGLE averages span half a percentage point, although the ordering changes with distance. Ground photographs give the highest of the three scores at $C=4$, and text gives the highest at $C=7$. At the intermediate distance $C=6$, success ranges from 85.2\% with ground photographs to 86.8\% with aerial goals. At $C=8$, the policy reaches the target in every evaluated episode with all three types of goal.

\begin{table}[t]
\caption{Ablation summary (average success ratio, \%). MM-GAG averages all three goal modalities. The w/o distillation variant uses student-only Stage~1. ViT-bigG supplies full-scene visual context in place of Qwen. The policy variants remove $U_t$ and its coverage bonus or remove $V_t$ during PPO and evaluation.}
\label{tab:ablation}
\centering
\small
\setlength{\tabcolsep}{3.0pt}
\begin{tabular}{lcccc}
\toprule
Variant & MASA & xBD & MM-GAG & Mean \\
\midrule
Full EAGLE & \textbf{78.1} & \textbf{76.4} & \textbf{76.5} & \textbf{77.0} \\
ViT teacher & 68.3 & 69.2 & 62.4 & 66.6 \\
w/o distillation & 59.1 & 59.2 & 58.8 & 59.0 \\
w/o uncertainty & 76.0 & 74.4 & 75.6 & 75.3 \\
w/o visitation & 77.9 & 76.1 & 75.6 & 76.5 \\
\bottomrule
\end{tabular}
\end{table}

For a given observation history, the student predicts an aerial feature for each cell before the goal is compared with the grid. Replacing an aerial crop with a description changes the goal embedding $\boldsymbol z_g$ and therefore the cell scores in $H_t$. The student can reuse its aerial features and variance estimates, and $V_t$ still records the cells seen along that path. The CNN can then use a peak in the heatmap to choose a move regardless of which encoder supplied the goal. If the two heatmaps lead to different moves, each path reveals its own patches and the tracker updates the maps accordingly.

\subsection{Ablation Studies}

\paragraph{Ablation of teacher-conditioned distillation}
The student-only variant keeps the student architecture, training masks, data, and optimization schedule. It trains the student to reconstruct masked aerial features and estimate their residual variance. The full model also trains a teacher with access to the complete scene and asks the student to match the teacher's feature predictions on those same masked cells. Each variant freezes its own student before PPO and trains the policy with the same settings.

Removing the teacher and its distillation loss lowers the three-dataset mean success ratio from 77.0\% to 59.0\% in Table~\ref{tab:ablation}. The drop is 19.0 percentage points on MASA, 17.2 on xBD, and 17.7 on MM-GAG. The effect thus appears both within the training domain and on new imagery. The connection to navigation is through $H_t$. Even before a cell has been visited, its predicted aerial feature can be compared with the goal. Of the variants in Table~\ref{tab:ablation}, removing teacher supervision produces the largest loss in navigation success on each dataset.

\paragraph{Choice of full-scene teacher features}
We replace the Qwen2.5-VL features supplied to the teacher with features extracted by ViT-bigG from the complete aerial scene. This comparison retains access to the entire search area during offline teaching. It changes the source of the teacher's context. ViT-bigG has a specific connection to the Qwen family. OpenCLIP ViT-bigG provided the initialization for the visual encoder in the original Qwen-VL~\cite{bai2023qwenvl}. The Qwen2.5-VL model used in the full system has a separately trained dynamic-resolution vision transformer~\cite{bai2025qwen25vl}. Its cached features are taken from the final hidden layer of the multimodal model.

With ViT-bigG context, success averages 68.3\% on MASA, 69.2\% on xBD, and 62.4\% on MM-GAG. The unweighted dataset mean is 66.6\%. Full-scene visual features recover 7.6 percentage points over the student-only mean of 59.0\%. Using Qwen features gives a further 10.4 points. The size of the difference depends on the benchmark. Qwen exceeds ViT-bigG by 9.8 points on MASA, 7.2 on xBD, and 14.1 on MM-GAG. The last value averages all three goal modalities. Both full-scene teacher variants score above the student-only model on each dataset.

\paragraph{Ablation of variance-based coverage}
Uncertainty plays two roles during PPO. For an unseen cell, $H_t$ measures agreement with the goal and $U_t$ estimates how much error to expect in the underlying aerial feature. A cell can score highly on both maps. Visiting it reveals the patch and sets its uncertainty entry to zero. The coverage bonus also accounts for changes in the estimated variance of other unseen cells as the student incorporates the new observation. Removing the uncertainty channel and the bonus together lowers mean success from 77.0\% to 75.3\%. The decrease is 2.1 points on MASA, 2.0 on xBD, and 0.9 on MM-GAG. These values measure the joint contribution of exposing uncertainty to the policy and using its reduction as a reward.

\paragraph{Ablation of the explicit visitation channel}
We remove the visitation channel $V_t$ from the policy input during PPO training and evaluation. The environment still uses the visit history to set $U_t$ to zero at visited cells. Mean success falls from 77.0\% to 76.5\%, with decreases of 0.2 percentage points on MASA, 0.3 on xBD, and 0.9 on MM-GAG.

\begin{table*}[!t]
\caption{Distance-wise success ratios for the student-only and policy ablations on aerial goals, in percent.}
\label{tab:distdetail}
\centering
\small
\setlength{\tabcolsep}{4.5pt}
\begin{tabular*}{\textwidth}{@{\extracolsep{\fill}}llrrrrrr@{}}
\toprule
Dataset & Configuration & $C=4$ & $C=5$ & $C=6$ & $C=7$ & $C=8$ & Avg. \\
\midrule
\multirow{4}{*}{MASA}
& Full EAGLE & 47.3 & 56.4 & 87.0 & 99.6 & 100.0 & 78.1 \\
& w/o distillation & 37.4 & 49.3 & 63.8 & 68.8 & 76.2 & 59.1 \\
& w/o uncertainty & 41.0 & 53.7 & 85.0 & 100.0 & 100.0 & 76.0 \\
& w/o visitation & 42.4 & 56.8 & 90.2 & 100.0 & 100.0 & 77.9 \\
\midrule
\multirow{4}{*}{xBD}
& Full EAGLE & 44.0 & 54.2 & 85.4 & 98.6 & 100.0 & 76.4 \\
& w/o distillation & 40.2 & 47.6 & 64.2 & 68.9 & 75.0 & 59.2 \\
& w/o uncertainty & 38.6 & 48.8 & 84.5 & 100.0 & 100.0 & 74.4 \\
& w/o visitation & 42.4 & 53.8 & 84.9 & 99.2 & 100.0 & 76.1 \\
\bottomrule
\end{tabular*}
\end{table*}

\subsection{Search Distance and Action Budget}
The nominal distance $C$ affects both the number of moves needed to reach a target and the number of possible destinations. To separate these two properties, consider the ordered start--goal pairs of a $5\times5$ grid. For an offset of $a$ rows and $b$ columns, the number of placements is $(5-|a|)(5-|b|)$. Summing over offsets at Manhattan distance $C$ gives
\begin{equation}
N_C=\sum_{\substack{|a|+|b|=C\\ |a|,|b|\leq4}}
     (5-|a|)(5-|b|).
\label{eq:pairs}
\end{equation}
There are 120 ordered pairs at distance four, 80 at five, 40 at six, 16 at seven, and four at eight. At distance eight each possible start is a corner and its destination is the opposite corner. Distance four admits a wider range of starting positions and multiple possible destinations for a given start. These counts describe the grid geometry and can be calculated without reference to any image.

The fixed budget of ten actions leaves six moves beyond a shortest route at distance four and two moves at distance eight. On this grid a step changes the Manhattan distance to a fixed goal by exactly one. One step away from the goal therefore costs two moves relative to a shortest route, since that extra distance must also be recovered. At distance eight there is room for one such detour. The policy faces a tighter movement budget at this distance but a much narrower distribution of goal positions. Success at eight cells and at four cells tests different aspects of the search.

Table~\ref{tab:distdetail} reports the distance profiles of the tracker and coverage ablations on aerial goals. On MASA, removing teacher distillation lowers success at $C=4$ by 9.9 points and at $C=7$ by 30.8 points. On xBD the corresponding differences are 3.8 and 29.7 points. Removing uncertainty has a different profile. At $C=4$, full EAGLE exceeds this variant by 6.3 points on MASA and 5.4 on xBD. At $C=8$ both configurations reach 100.0\% on both datasets. The dataset means therefore combine differences at short distances with settings in which the two systems already attain the maximum score.

\section{Discussion}
\paragraph{What a new observation changes}
Entering a cell affects the policy input in two ways. First, it replaces the blank patch at that coordinate with RGB evidence that the tracker can use at all locations. Second, it removes the visited cell from the uncertainty sum in the coverage reward. The latter operation depends only on the visit and the prior variance at the cell. The former can increase or decrease uncertainty elsewhere. A transition that reveals an unexpected boundary between a road and vegetation, for example, could revise predictions on either side. The reward uses the net reduction summed over the grid, while the next policy decision retains the location of every change.

\paragraph{Revisits and the search budget}
The agent can revisit a cell to reach another part of the grid. Each revisit uses one move, incurs the step cost, and earns no coverage bonus.

\paragraph{How coverage is credited}
Suppose the agent moves into a previously unseen cell $j$. Let $S_t$ denote the cells still unobserved after this move and let $v_{t,i}$ denote the student's raw variance estimate for cell $i$. The change in uncertainty before applying the positive-part operator can be written as
\begin{equation}
 \lVert U_{t-1}\rVert_1-\lVert U_t\rVert_1
 =v_{t-1,j}+\sum_{i\in S_t}\big[v_{t-1,i}-v_{t,i}\big].
\label{eq:coverage-change}
\end{equation}
The first term is present for every newly observed cell. The second depends on how that observation changes the rest of the prediction. If all other estimates stay fixed, observing a cell with a large variance earns more coverage reward than observing one whose feature was already predictable. If the new observations raise uncertainty elsewhere, that increase offsets the first term and may reduce the bonus to zero. The heatmap bonus can still reward this move if the newly occupied location provides stronger goal evidence than any earlier location. The two rewards thus give different credit to a single observation.

\paragraph{Cross-modal behavior of the policy channels}
The joint uncertainty ablation and the visitation ablation leave different information available to the policy. With the uncertainty map and its bonus removed, mean MM-GAG success is 75.9\% for aerial queries, 75.7\% for ground photographs, and 75.2\% for text. When the separate visitation channel is removed, these values are 76.1\%, 75.8\%, and 74.8\%. For the text query the latter difference from the full policy is 1.6 percentage points. For the ground query the difference is 0.5 points. Even though opening a new cell uses the same update rule in all three cases, the map of similarities differs across queries. Decisions that depend on covered regions can consequently affect the three modalities differently. These values summarize outcomes over paths chosen by each policy.

\paragraph{Implications for an image-based navigation system}
During training, each complete scene is encoded once and reused across partial views. During navigation, the student updates its grid after each observation and compares it with a goal embedding computed once per episode. Ground photographs and text use different encoders to query the same grid through a shared 512-dimensional representation.

The tracker assumes that each observation can be placed at a known location on a fixed-scale aerial grid. In continuous flight, changes in altitude and viewing angle would require successive onboard images to be registered to a common map. Flight tests could assess how registration errors affect target search.

\section{Conclusion}

We introduced EAGLE, an active geo-localization agent that combines spatial belief tracking with offline vision-language distillation. From the aerial patches observed along its route, EAGLE predicts features across the search grid, including at locations it has yet to visit. The policy uses the similarity of these features to the goal, their estimated uncertainty, and the visit history to choose its next move. A policy trained with aerial goals on MASA transferred to disaster imagery on xBD and located targets specified by ground-level photographs or text on MM-GAG. The gains over student-only training across all three benchmarks show that supervision from complete scenes can guide exploration when only local views are available during navigation.

\section*{Acknowledgments}

We thank Mi et al.~\cite{mi2025geoexplorer} for making the GeoExplorer code used in this work publicly available. We used OpenAI Codex to help write and polish passages throughout the manuscript and to revise the research code. All AI-assisted code changes underwent human review.

\AddToHook{cmd/thebibliography/after}{\interlinepenalty=10000\relax}
\IEEEtriggeratref{36}
\bibliographystyle{IEEEtran}
\bibliography{references}

\begin{thebibliography}{10}
\providecommand{\url}[1]{#1}
\csname url@samestyle\endcsname
\providecommand{\newblock}{\relax}
\providecommand{\bibinfo}[2]{#2}
\providecommand{\BIBentrySTDinterwordspacing}{\spaceskip=0pt\relax}
\providecommand{\BIBentryALTinterwordstretchfactor}{4}
\providecommand{\BIBentryALTinterwordspacing}{\spaceskip=\fontdimen2\font plus
\BIBentryALTinterwordstretchfactor\fontdimen3\font minus
  \fontdimen4\font\relax}
\providecommand{\BIBforeignlanguage}[2]{{%
\expandafter\ifx\csname l@#1\endcsname\relax
\typeout{** WARNING: IEEEtran.bst: No hyphenation pattern has been}%
\typeout{** loaded for the language `#1'. Using the pattern for}%
\typeout{** the default language instead.}%
\else
\language=\csname l@#1\endcsname
\fi
#2}}
\providecommand{\BIBdecl}{\relax}
\BIBdecl

\bibitem{sarkar2024gomaa}
A.~Sarkar, S.~Sastry, A.~Pirinen, C.~Zhang, N.~Jacobs, and Y.~Vorobeychik,
  ``Gomaa-geo: Goal modality agnostic active geo-localization,'' \emph{Advances
  in Neural Information Processing Systems}, vol.~37, pp. 104\,934--104\,964,
  2024.

\bibitem{mi2025geoexplorer}
L.~Mi, M.~B{\'e}chaz, Z.~Chen, A.~Bosselut, and D.~Tuia, ``Geoexplorer: Active
  geo-localization with curiosity-driven exploration,'' in \emph{Proceedings of
  the IEEE/CVF International Conference on Computer Vision}, 2025, pp.
  6122--6131.

\bibitem{workman2015wide}
S.~Workman, R.~Souvenir, and N.~Jacobs, ``Wide-area image geolocalization with
  aerial reference imagery,'' in \emph{Proceedings of the IEEE International
  Conference on Computer Vision}, 2015, pp. 3961--3969.

\bibitem{zhu2021vigor}
S.~Zhu, T.~Yang, and C.~Chen, ``Vigor: Cross-view image geo-localization beyond
  one-to-one retrieval,'' in \emph{Proceedings of the IEEE/CVF Conference on
  Computer Vision and Pattern Recognition}, 2021, pp. 3640--3649.

\bibitem{berton2022cosplace}
G.~Berton, C.~Masone, and B.~Caputo, ``Rethinking visual geo-localization for
  large-scale applications,'' in \emph{Proceedings of the IEEE/CVF Conference
  on Computer Vision and Pattern Recognition}, 2022, pp. 4878--4888.

\bibitem{berton2022deep}
G.~Berton, R.~Mereu, G.~Trivigno, C.~Masone, G.~Csurka, T.~Sattler, and
  B.~Caputo, ``Deep visual geo-localization benchmark,'' in \emph{Proceedings
  of the IEEE/CVF Conference on Computer Vision and Pattern Recognition}, 2022,
  pp. 5396--5407.

\bibitem{vivanco2023geoclip}
V.~Vivanco~Cepeda, G.~K. Nayak, and M.~Shah, ``Geoclip: Clip-inspired alignment
  between locations and images for effective worldwide geo-localization,''
  \emph{Advances in Neural Information Processing Systems}, vol.~36, pp.
  8690--8701, 2023.

\bibitem{mi2024congeo}
L.~Mi, C.~Xu, J.~Castillo-Navarro, S.~Montariol, W.~Yang, A.~Bosselut, and
  D.~Tuia, ``Congeo: Robust cross-view geo-localization across ground view
  variations,'' in \emph{European Conference on Computer Vision}.\hskip 1em
  plus 0.5em minus 0.4em\relax Springer, 2024, pp. 214--230.

\bibitem{astruc2024osv5m}
G.~Astruc, N.~Dufour, I.~Siglidis, C.~Aronssohn, N.~Bouia, S.~Fu, R.~Loiseau,
  V.~N. Nguyen, C.~Raude, E.~Vincent \emph{et~al.}, ``Openstreetview-5m: The
  many roads to global visual geolocation,'' in \emph{Proceedings of the
  IEEE/CVF Conference on Computer Vision and Pattern Recognition}, 2024, pp.
  21\,967--21\,977.

\bibitem{fervers2023uncertainty}
F.~Fervers, S.~Bullinger, C.~Bodensteiner, M.~Arens, and R.~Stiefelhagen,
  ``Uncertainty-aware vision-based metric cross-view geolocalization,'' in
  \emph{Proceedings of the IEEE/CVF Conference on Computer Vision and Pattern
  Recognition}, 2023, pp. 21\,621--21\,631.

\bibitem{radford2021clip}
A.~Radford, J.~W. Kim, C.~Hallacy, A.~Ramesh, G.~Goh, S.~Agarwal, G.~Sastry,
  A.~Askell, P.~Mishkin, J.~Clark \emph{et~al.}, ``Learning transferable visual
  models from natural language supervision,'' in \emph{Proceedings of the 38th
  International Conference on Machine Learning}.\hskip 1em plus 0.5em minus
  0.4em\relax PMLR, 2021, pp. 8748--8763.

\bibitem{dhakal2024sat2cap}
A.~Dhakal, A.~Ahmad, S.~Khanal, S.~Sastry, H.~Kerner, and N.~Jacobs,
  ``{Sat2Cap}: Mapping fine-grained textual descriptions from satellite
  images,'' in \emph{Proceedings of the IEEE/CVF Conference on Computer Vision
  and Pattern Recognition Workshops}, 2024, pp. 533--542.

\bibitem{chen2026dide}
Z.~Chen, T.~Li, D.~Li, J.~Gou, C.~Wang, K.~Gao, J.~Li, and Z.~Gong,
  ``Leveraging multi-view images to learn domain-invariant discriminative
  embeddings for cross-view geo-localization,'' \emph{IEEE Transactions on
  Circuits and Systems for Video Technology}, vol.~36, no.~3, pp. 2669--2684,
  2026.

\bibitem{xu2026aerial}
H.~Xu, Z.~Liu, Y.~Luomei, and F.~Xu, ``Aerial vision-language navigation with a
  unified framework for spatial, temporal and embodied reasoning,'' \emph{IEEE
  Transactions on Circuits and Systems for Video Technology}, vol.~36, no.~10,
  pp. 14\,942--14\,957, 2026.

\bibitem{bai2025qwen25vl}
\BIBentryALTinterwordspacing
S.~Bai, K.~Chen, X.~Liu, J.~Wang, W.~Ge, S.~Song, K.~Dang, P.~Wang, S.~Wang,
  J.~Tang, H.~Zhong, Y.~Zhu, M.~Yang, Z.~Li, J.~Wan, P.~Wang, W.~Ding, Z.~Fu,
  Y.~Xu, J.~Ye, X.~Zhang, T.~Xie, Z.~Cheng, H.~Zhang, Z.~Yang, H.~Xu, and
  J.~Lin, ``Qwen2.5-vl technical report,'' 2025. [Online]. Available:
  \url{https://arxiv.org/abs/2502.13923}
\BIBentrySTDinterwordspacing

\bibitem{liu2023llava}
H.~Liu, C.~Li, Q.~Wu, and Y.~J. Lee, ``Visual instruction tuning,''
  \emph{Advances in neural information processing systems}, vol.~36, pp.
  34\,892--34\,916, 2023.

\bibitem{chen2024internvl}
Z.~Chen, J.~Wu, W.~Wang, W.~Su, G.~Chen, S.~Xing, M.~Zhong, Q.~Zhang, X.~Zhu,
  L.~Lu \emph{et~al.}, ``Internvl: Scaling up vision foundation models and
  aligning for generic visual-linguistic tasks,'' in \emph{Proceedings of the
  IEEE/CVF Conference on Computer Vision and Pattern Recognition}, 2024, pp.
  24\,185--24\,198.

\bibitem{hinton2015distilling}
G.~Hinton, O.~Vinyals, and J.~Dean, ``Distilling the knowledge in a neural
  network,'' \emph{arXiv preprint arXiv:1503.02531}, 2015.

\bibitem{fang2021distillvlm}
Z.~Fang, J.~Wang, X.~Hu, L.~Wang, Y.~Yang, and Z.~Liu, ``Compressing
  visual-linguistic model via knowledge distillation,'' in \emph{Proceedings of
  the IEEE/CVF International Conference on Computer Vision}, 2021, pp.
  1428--1438.

\bibitem{wu2023tinyclip}
K.~Wu, H.~Peng, Z.~Zhou, B.~Xiao, M.~Liu, L.~Yuan, H.~Xuan, M.~Valenzuela,
  X.~S. Chen, X.~Wang \emph{et~al.}, ``Tinyclip: Clip distillation via affinity
  mimicking and weight inheritance,'' in \emph{Proceedings of the IEEE/CVF
  International Conference on Computer Vision}, 2023, pp. 21\,970--21\,980.

\bibitem{wang2023efficientvlm}
T.~Wang, W.~Zhou, Y.~Zeng, and X.~Zhang, ``Efficientvlm: Fast and accurate
  vision-language models via knowledge distillation and modal-adaptive
  pruning,'' in \emph{Findings of the Association for Computational
  Linguistics: ACL 2023}, 2023, pp. 13\,899--13\,913.

\bibitem{pirinen2022aerial}
A.~Pirinen, A.~Samuelsson, J.~Backsund, and K.~{\AA}str{\"o}m, ``Aerial view
  localization with reinforcement learning: Towards emulating
  search-and-rescue,'' \emph{arXiv preprint arXiv:2209.03694}, 2022.

\bibitem{chen2021decision}
L.~Chen, K.~Lu, A.~Rajeswaran, K.~Lee, A.~Grover, M.~Laskin, P.~Abbeel,
  A.~Srinivas, and I.~Mordatch, ``Decision transformer: Reinforcement learning
  via sequence modeling,'' \emph{Advances in neural information processing
  systems}, vol.~34, pp. 15\,084--15\,097, 2021.

\bibitem{gu2022vlnsurvey}
J.~Gu, E.~Stefani, Q.~Wu, J.~Thomason, and X.~E. Wang, ``Vision-and-language
  navigation: A survey of tasks, methods, and future directions,'' in
  \emph{Proceedings of the 60th Annual Meeting of the Association for
  Computational Linguistics (Volume 1: Long Papers)}, 2022, pp. 7606--7623.

\bibitem{liu2023aerialvln}
S.~Liu, H.~Zhang, Y.~Qi, P.~Wang, Y.~Zhang, and Q.~Wu, ``Aerialvln:
  Vision-and-language navigation for uavs,'' in \emph{Proceedings of the
  IEEE/CVF International Conference on Computer Vision}, 2023, pp.
  15\,384--15\,394.

\bibitem{shah2023vint}
D.~Shah, A.~Sridhar, N.~Dashora, K.~Stachowicz, K.~Black, N.~Hirose, and
  S.~Levine, ``Vint: A foundation model for visual navigation,'' \emph{arXiv
  preprint arXiv:2306.14846}, 2023.

\bibitem{kaelbling1998pomdp}
L.~P. Kaelbling, M.~L. Littman, and A.~R. Cassandra, ``Planning and acting in
  partially observable stochastic domains,'' \emph{Artificial intelligence},
  vol. 101, no. 1-2, pp. 99--134, 1998.

\bibitem{thrun2002probabilistic}
S.~Thrun, ``Probabilistic robotics,'' \emph{Communications of the ACM},
  vol.~45, no.~3, pp. 52--57, 2002.

\bibitem{gupta2017cognitive}
S.~Gupta, J.~Davidson, S.~Levine, R.~Sukthankar, and J.~Malik, ``Cognitive
  mapping and planning for visual navigation,'' in \emph{Proceedings of the
  IEEE conference on computer vision and pattern recognition}, 2017, pp.
  2616--2625.

\bibitem{chaplot2020activeslam}
D.~S. Chaplot, D.~Gandhi, S.~Gupta, A.~Gupta, and R.~Salakhutdinov, ``Learning
  to explore using active neural slam,'' \emph{arXiv preprint
  arXiv:2004.05155}, 2020.

\bibitem{stadie2015incentivizing}
B.~C. Stadie, S.~Levine, and P.~Abbeel, ``Incentivizing exploration in
  reinforcement learning with deep predictive models,'' \emph{arXiv preprint
  arXiv:1507.00814}, 2015.

\bibitem{pathak2017curiosity}
D.~Pathak, P.~Agrawal, A.~A. Efros, and T.~Darrell, ``Curiosity-driven
  exploration by self-supervised prediction,'' in \emph{International
  conference on machine learning}.\hskip 1em plus 0.5em minus 0.4em\relax PMLR,
  2017, pp. 2778--2787.

\bibitem{burda2019largescale}
\BIBentryALTinterwordspacing
Y.~Burda, H.~Edwards, D.~Pathak, A.~Storkey, T.~Darrell, and A.~A. Efros,
  ``Large-scale study of curiosity-driven learning,'' in \emph{International
  Conference on Learning Representations}, 2019. [Online]. Available:
  \url{https://openreview.net/forum?id=rJNwDjAqYX}
\BIBentrySTDinterwordspacing

\bibitem{burda2019exploration}
\BIBentryALTinterwordspacing
Y.~Burda, H.~Edwards, A.~Storkey, and O.~Klimov, ``Exploration by random
  network distillation,'' in \emph{International Conference on Learning
  Representations}, 2019. [Online]. Available:
  \url{https://openreview.net/forum?id=H1lJJnR5Ym}
\BIBentrySTDinterwordspacing

\bibitem{kim2020progress}
K.~Kim, M.~Sano, J.~De~Freitas, N.~Haber, and D.~Yamins, ``Active world model
  learning with progress curiosity,'' in \emph{International conference on
  machine learning}.\hskip 1em plus 0.5em minus 0.4em\relax PMLR, 2020, pp.
  5306--5315.

\bibitem{sun2026dyncur}
Y.~Sun, Y.~Zhang, and P.~Zhu, ``{DynCur-Geo}: Dynamic curiosity reward shaping
  for multimodal active geo-localization,'' \emph{arXiv preprint
  arXiv:2608.18673}, 2026.

\bibitem{nix1994variance}
D.~A. Nix and A.~S. Weigend, ``Estimating the mean and variance of the target
  probability distribution,'' in \emph{Proceedings of 1994 ieee international
  conference on neural networks (ICNN'94)}, vol.~1.\hskip 1em plus 0.5em minus
  0.4em\relax IEEE, 1994, pp. 55--60.

\bibitem{georgakis2022}
G.~Georgakis, B.~Bucher, K.~Schmeckpeper, S.~Singh, and K.~Daniilidis,
  ``Learning to map for active semantic goal navigation,'' in
  \emph{International Conference on Learning Representations}, 2022.

\bibitem{liu2021swin}
Z.~Liu, Y.~Lin, Y.~Cao, H.~Hu, Y.~Wei, Z.~Zhang, S.~Lin, and B.~Guo, ``Swin
  transformer: Hierarchical vision transformer using shifted windows,'' in
  \emph{Proceedings of the IEEE/CVF international conference on computer
  vision}, 2021, pp. 10\,012--10\,022.

\bibitem{schulman2017ppo}
J.~Schulman, F.~Wolski, P.~Dhariwal, A.~Radford, and O.~Klimov, ``Proximal
  policy optimization algorithms,'' \emph{arXiv preprint arXiv:1707.06347},
  2017.

\bibitem{schulman2015high}
J.~Schulman, P.~Moritz, S.~Levine, M.~Jordan, and P.~Abbeel, ``High-dimensional
  continuous control using generalized advantage estimation,'' \emph{arXiv
  preprint arXiv:1506.02438}, 2015.

\bibitem{gupta2019xbd}
R.~Gupta, B.~Goodman, N.~Patel, R.~Hosfelt, S.~Sajeev, E.~Heim, J.~Doshi,
  K.~Lucas, H.~Choset, and M.~Gaston, ``Creating xbd: A dataset for assessing
  building damage from satellite imagery,'' in \emph{Proceedings of the
  IEEE/CVF conference on computer vision and pattern recognition workshops},
  2019, pp. 10--17.

\bibitem{bai2023qwenvl}
J.~Bai, S.~Bai, S.~Yang, S.~Wang, S.~Tan, P.~Wang, J.~Lin, C.~Zhou, and
  J.~Zhou, ``{Qwen-VL}: A versatile vision-language model for understanding,
  localization, text reading, and beyond,'' \emph{arXiv preprint
  arXiv:2308.12966}, 2023.

\end{thebibliography}

\end{document}